\pdfoutput=1

\documentclass[11pt]{article}

\usepackage{acl}

\usepackage{times}
\usepackage{latexsym}
\usepackage{amsfonts}
\usepackage[acronym]{glossaries}
\glsdisablehyper
\usepackage{graphicx}
\graphicspath{ {./images/} }
\DeclareGraphicsExtensions{.pdf,.png}
\usepackage{caption}
\usepackage{subcaption}
\usepackage[T1]{fontenc}

\usepackage[utf8]{inputenc}

\usepackage{microtype,tabularx}

\usepackage{booktabs}
\usepackage{multirow}
\usepackage{hyperref}

\title{Leveraging knowledge graphs to update scientific word embeddings using latent semantic imputation}

\author{Jason Hoelscher-Obermaier\thanks{\ \ Co-first authors}, Edward Stevinson\footnotemark[1] \\ 
{\bf Valentin Stauber, Ivaylo Zhelev} \\ {\bf Victor Botev\thanks{\ \ Co-PIs}, Ronin Wu\footnotemark[2], Jeremy Minton\footnotemark[2]} \\
Iris AI, Bekkestua, Norway \\
\texttt{jason@iris.ai}
}

\begin{document}
\newcommand\semantic[1]{\mathbf{#1}_s}
\newcommand\semanticembedding[0]{E_s}
\newcommand\semanticspace[0]{\mathbb{R}^s}
\newcommand\domain[1]{\mathbf{#1}_d}
\newcommand{\domainembedding}[0]{E_d}
\newcommand{\domainspace}[0]{\mathbb{R}^d}

\newacronym{lsi}{LSI}{latent semantic imputation}
\newacronym{oov}{OOV}{out of vocabulary}
\newacronym{mesh}{MeSH}{Medical Subject Headings}
\newacronym{knnmst}{kNN-MST}{k-nearest-neighbors minimum-spanning-tree}
\newacronym{nlp}{NLP}{natural language processing}
\maketitle
\begin{abstract}
The most interesting words in scientific texts will often be novel or rare. This presents a challenge for scientific word embedding models to determine quality embedding vectors for useful terms that are infrequent or newly emerging. We demonstrate how \gls{lsi} can address this problem by imputing embeddings for domain-specific words from up-to-date knowledge graphs while otherwise preserving the original word embedding model. 
We use the \gls{mesh} knowledge graph to impute embedding vectors for biomedical terminology without retraining and evaluate the resulting embedding model on a domain-specific word-pair similarity task.
We show that \gls{lsi} can produce reliable embedding vectors for rare and \gls{oov} terms in the biomedical domain.
\end{abstract}
\glsresetall

\section{Introduction}
Word embeddings are powerful representations of the semantic and syntactic properties of words that facilitate high performance in \gls{nlp} tasks.
Because these models completely rely on a training corpus, they can struggle to reliably represent words which are infrequent, or missing entirely, in that corpus.
The latter will happen for any new terminology emerging after training is complete.

Rapid emergence of new terminology and a long tail of highly significant but rare words are characteristic of technical domains, but these terms are often of particular importance to \gls{nlp} tasks within these domains.
This drives a need for methods to generate reliable embeddings of rare and novel words.
At the same time, there are efforts in many scientific fields to construct large, highly specific and continuously updated knowledge graphs that capture information about these exact terms.
Can we leverage these knowledge graphs to mitigate the short-comings of word embeddings on rare, novel and domain-specific words?

We investigate one method for achieving this information transfer, \gls{lsi} \citep{yaoEnhancingDomainWord2019}. In \gls{lsi} the embedding vector for a given word, $w$, is imputed as a weighted average of existing embedding vectors, where the weights are inferred from the local neighborhood structure of a corresponding embedding vector, $\domain{w}$, in a domain-specific embedding space. We study how to apply \gls{lsi} in the context of the biomedical domain using the \gls{mesh} knowledge graph \citep{lipscombMedicalSubjectHeadings2000}, but expect the methodology to be applicable to other scientific domains.

\section{Related work}

\textbf{Embeddings for rare/\gls{oov} words.}
Early methods for embedding rare words relied on explicitly provided morphological information \citep{alexandrescuFactoredNeuralLanguage2006, sakMorphologybasedSubwordLanguage2010, lazaridouCompositionallyDerivedRepresentations2013, bothaCompositionalMorphologyWord2014, luongAchievingOpenVocabulary2016, qiuColearningWordRepresentations2014}. 
More recent approaches avoid dependence on explicit morphological information by learning representations for fixed-length character n-grams that do not have a direct linguistic interpretation \citep{bojanowskiEnrichingWordVectors2017, zhaoGeneralizingWordEmbeddings2018}. 
Alternatively, the subword structure used for generalization beyond a fixed vocabulary can be learnt from data using techniques such as byte-pair encoding \citep{sennrichNeuralMachineTranslation2016, gage1994DataCompression} or the WordPiece algorithm \citep{chusterVoiceSearch2012}. Embeddings for arbitrary strings can also be generated using character-level recurrent networks \citep{lingFindingFunctionForm2015, xieRepresentationLearningKnowledge2016, pinterMimickingWordEmbeddings2017}. 
These approaches, as well as transformer-based methods mentioned below, provide some \gls{oov} generalization capability but are unlikely to be a general solution since they will struggle with novel terms whose meaning is not implicit in the subword structure such as, \textit{e.g.}, eponyms. Note that we experimented with fastText and it performed worse than our approach.

\textbf{Word embeddings for the biomedical domain.}
Much research has focused on how to best generate biomedical-specific embeddings and provide models to improve performance on downstream \gls{nlp} tasks \citep{majorUtilityGeneralSpecific2018, pyysaloDistributionalSemanticsResources2013, chiuHowTrainGood2016, zhangBioWordVecImprovingBiomedical2019}. Work in the biomedical domain has investigated optimal hyperparameters for embedding training \cite{chiuHowTrainGood2016}, the influence of the training corpus \citep{pakhomovCorpusDomainEffects2016, wangComparisonWordEmbeddings2018, laiHowGenerateGood2016}, and the advantage of subword-based embeddings \citep{zhangBioWordVecImprovingBiomedical2019}. Word embeddings for clinical applications have been proposed \citep{ghoshCharacterisingDiseases2016, fanDietryEmbeddings2019} and an overview was provided in \citet{kalyanSECNLPSurveyEmbeddings2020}. More recently, transformer models have been successfully adapted to the biomedical domain yielding contextual, domain-specific embedding models \citep{pengTransferLearningBiomedical2019, leeBioBERTPretrainedBiomedical2019,beltagySciBERTPretrainedLanguage2019, phanSciFiveTexttotextTransformer2021}. Whilst these works highlight the benefits of domain-specific training corpora this class of approaches requires retraining to address the \gls{oov} problem.

\textbf{Improving word embeddings using domain information.}
Our problem task requires improving a provided embedding model for a given domain, without detrimental effects on other domains.

\citet{zhangBioWordVecImprovingBiomedical2019} use random walks over the \gls{mesh} headings knowledge graph to generate additional training text to be used during the word embedding training.
Similar ideas have led to using regularization terms that leverage an existing embedding during training of a new embedding to preserve information from an original embedding during training on a new corpus \citep{yangSimpleRegularizationbasedAlgorithm2017}.
Of course, these methods require the complete training of one or more embedding models.

\citet{faruquiRetrofittingWordVectors2014} achieve a similar result more efficiently by defining a convex objective function that balances preserving an existing embedding with decreasing the distance between related vectors, based on external data sources such as a lexicon. This technique has been applied in the biomedical domain \citep{yuRetrofittingWordVectors2016, yuRetrofittingConceptVector2017}, but has limited ability to infer new vocabulary because without the contribution from the original embedding this reduces to an average of related vectors.

Another approach is to extend the embedding dimension to create space for encoding new information.
This can be as simple as vector concatenation from another embedding \citep{yangSimpleRegularizationbasedAlgorithm2017}, possibly followed by dimensionality reduction \citep{shalabyWordEmbeddingsLearning2018}.
Alternatively, new dimensions can be derived from existing vectors based on external information like synonym pairs \citep{joExtrofittingEnrichingWord2018}.
Again, this has limited ability to infer new vocabulary.

All of these methods change the original embedding, which limits applicability in use-cases where the original embedding quality must be retained or where incremental updates from many domains are required.
The optimal alignment of two partially overlapping word embedding spaces has been studied in the literature on multilingual word embeddings \citep{nakasholeKnowledgeDistillationBilingual2017, jawanpuriaLearningMultilingualWord2019, alauxUnsupervisedHyperalignmentMultilingual2019} and provides a mechanism to patch an existing embedding with information from a domain-specific embedding. Unfortunately, it assumes the embedding spaces have the same structure, meaning it is not suitable when the two embeddings encode different types of information, such as semantic information from text and relational information from a knowledge base.

\section{Latent Semantic Imputation}
\Gls{lsi}, the approach pursued in this paper, represents embedding vectors for new words as weighted averages over existing word embedding vectors with the weights derived from a domain-specific feature matrix \citep{yaoEnhancingDomainWord2019}. This process draws insights from Locally Linear Embedding \citep{roweisLocallyLinear2000}. Specifically, a local neighborhood in a high-dimensional word embedding space $\semanticembedding$ ($s$ for semantic) can be approximated by a lower-dimensional manifold embedded in that space. Hence, an embedding vector $\semantic{w}$ for a word $w$ in that local neighborhood can be approximated as a weighted average over a small number of neighboring vectors.

\begin{figure}[h]
\centering
\includegraphics[width=7.75cm, height=11.3cm]{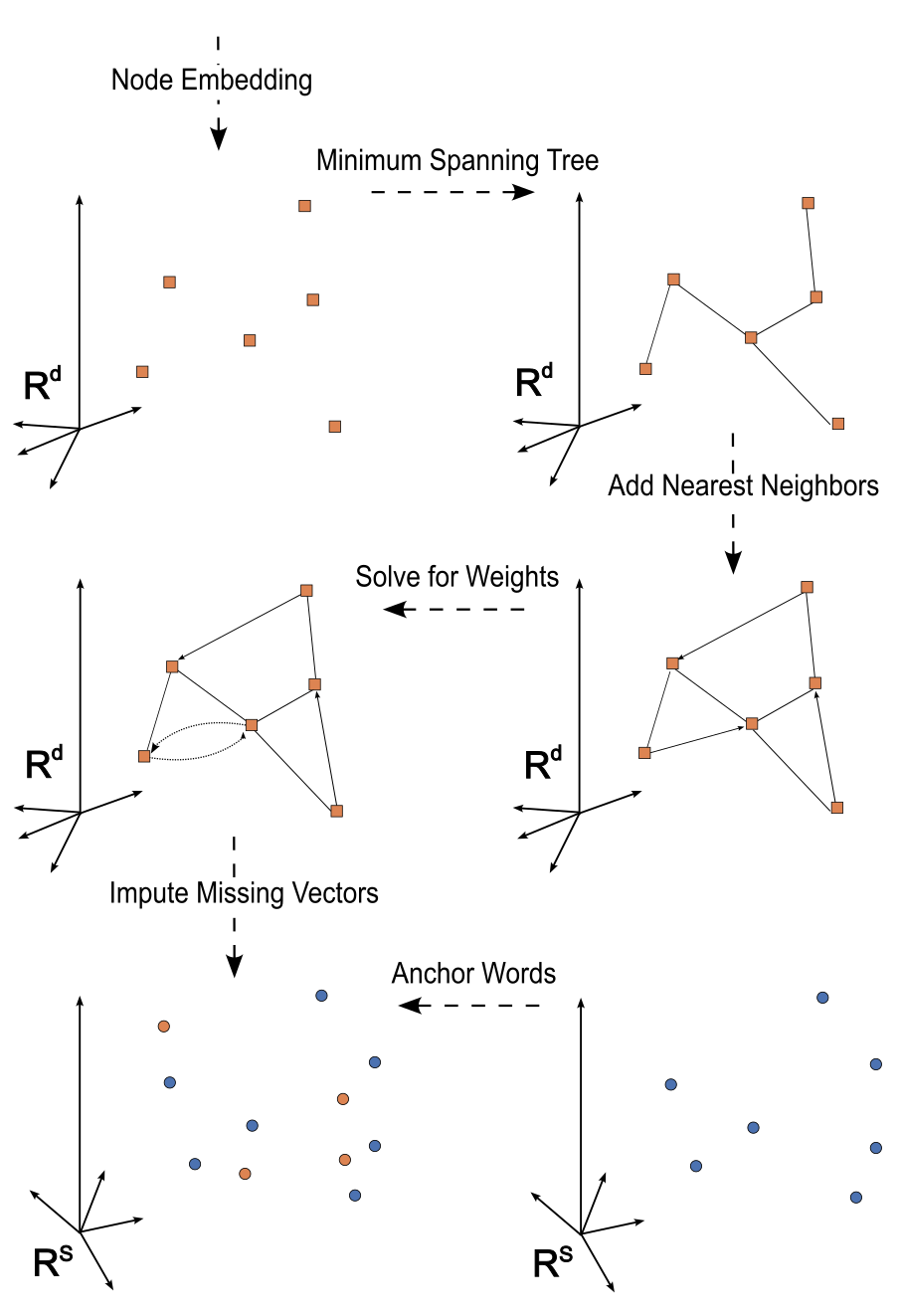}
\vspace*{-3mm}
\caption{Latent Semantic Imputation. $\domainspace$ is the domain space and $\semanticspace$ is the semantic space.}
\label{fig:lsiMethod}
\end{figure}

This would be useful to construct a vector of a new word $w$ if we could determine the weights for the average over neighboring terms. But since, by assumption, we do not know $w$'s word embedding vector $\semantic{w}$, we also do not know its neighborhood in $\semanticembedding$. The main insight of \gls{lsi} is that we can use the local neighborhood of $w$'s embedding $\domain{w}$ in a domain-specific space, $\domainembedding$, as a proxy for that neighborhood in the semantic space of our word-embedding model, $\semanticembedding$. The weights used for constructing an embedding for $w$ in $\semanticembedding$ are calculated from the domain space as shown in Fig.~\ref{fig:lsiMethod}: a \gls{knnmst} is built from the domain space features. Then the L2-distance between $\domain{w}$ and a weighted average over its neighbors in the \gls{knnmst} is minimized using non-negative least squares. The resulting weights are used to impute the missing embedding vectors in $\semanticembedding$ using the power iteration method. This procedure crucially relies on the existence of words with good representations in both $\semanticembedding$ and $\domainembedding$, referred to as anchor terms, which serve as data from which the positions of the derived embedding vectors are constructed.

\section{Methodology}
We extend the original \gls{lsi} procedure described above in a few key ways.
Instead of using a numeric data matrix as the domain data source of \gls{lsi}, we use a node embedding model trained on a domain-specific knowledge graph to obtain $\domainembedding$.
As knowledge graphs are used as a source of structured information in many fields, we expect our method to be applicable to many scientific domains. Knowledge graphs are prevalent in scientific fields as they serve as a means to organise and store scientific data, as well as to aid downstream tasks such as reasoning and exploration. Their structure and ability to represent different relationship types makes it relatively easy to integrate new data, meaning they can evolve to reflect changes in a field and as new data becomes available.

We use the 2021 RDF dump of the \gls{mesh} knowledge graph (available at \url{https://id.nlm.nih.gov/mesh/}). The complete graph consists of 2,327,188 nodes and 4,272,681 edges, which we reduce into a simpler, smaller, and undirected graph to be fed into a node embedding algorithm. We extract a subgraph consisting of solely the nodes of type "ns0\_\_TopicalDescriptor" and the nodes of type "ns0\_\_Concept" that are directly connected to the topical descriptors via any relationship type. The relationship types and directionality were removed. This results in 58,695 nodes and 113,094 edges. 

We use the node2vec graph embedding algorithm \citep{grover2016} on this subgraph to produce an embedding matrix of 58,695 vectors with dimension 200 (orange squares in Fig.~\ref{fig:our_methodology}). The hyperparameters are given in Appendix \ref{sec:app:hyperparams_mesh_n2v}. These node embeddings form the domain-specific space, $\domainembedding$, as described in the previous section. We note that in preliminary experiments, the adjacency matrix of the knowledge graph was used directly as $\domainembedding$ but this yielded imputed embeddings that performed poorly.

To provide the mapping between the \gls{mesh} nodes and the word embedding vocabulary we normalize the human-readable "rdfs\_\_label" node property by replacing spaces with hyphens and lower-casing. The anchor terms are then identified as the normalized words that match between the graph labels and the vocabulary of the word-embedding model; resulting in 12,676 anchor terms. As an example, "alpha-2-hs-glycoprotein" appears as both a node in the reduced graph and in the word-embedding model, along with its neighbors in the \gls{knnmst}, which include "neoglycoproteins" and "alpha-2-antiplasmin". These serve to stabilise the positions of unknown word embedding vectors for domain space nodes which did not have corresponding representations in the semantic space during the \gls{lsi} procedure.

\Gls{lsi} has one key hyper-parameter: the minimal degree of the \gls{knnmst} graph, $k$.
The stopping criterion of the power iteration method is controlled by another parameter, $\eta$, but any sufficiently small value should allow adequate convergence and have minimal impact on the resulting vectors.
Following \citet{yaoEnhancingDomainWord2019} we set $\eta=10^{-4}$ but we use a larger $k=50$ since initial experiments showed a better performance for larger values of $k$.

\begin{figure}
\centering
\includegraphics[width=7.75cm, height=9.51cm]{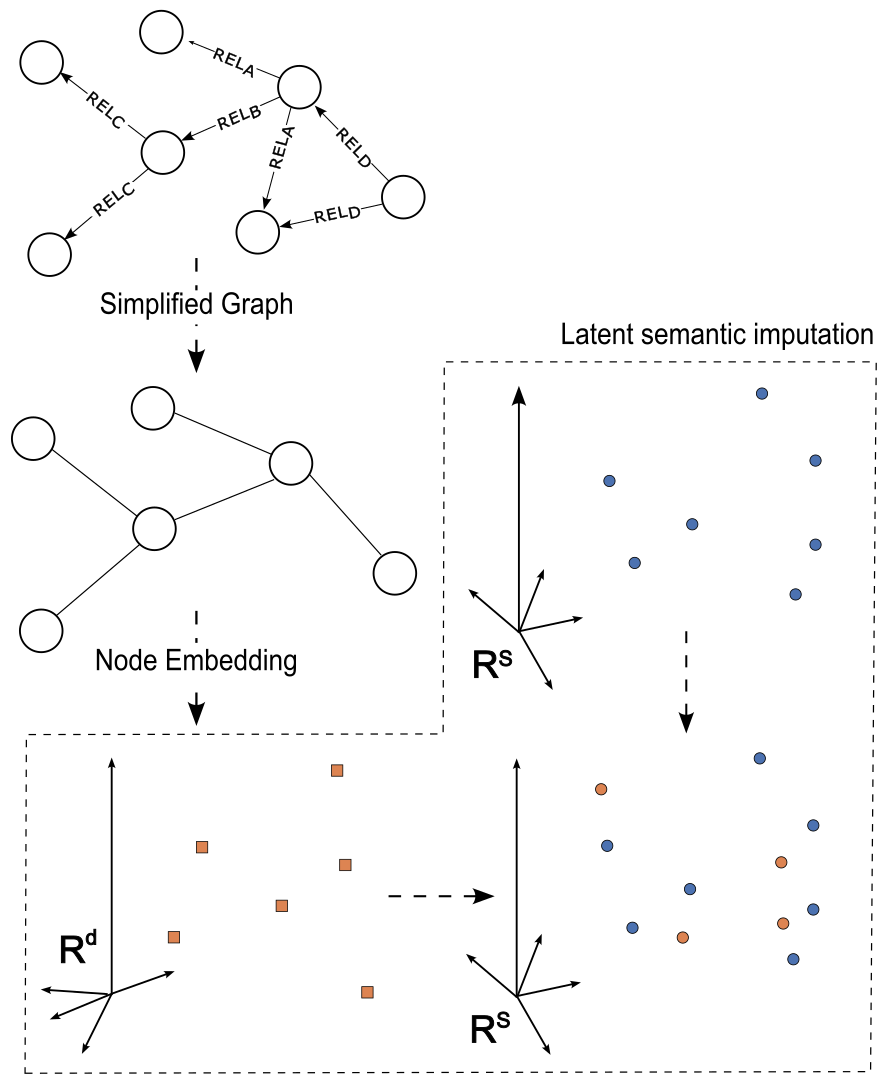}
\vspace*{-3mm}
\caption{Extended latent semantic imputation pipeline. A knowledge graph is simplified to a smaller, undirected graph. This is used to derive the node embedding model used in LSI (see Fig. {\ref{fig:lsiMethod}}) to impute missing terms in the semantic space.}
\label{fig:our_methodology}
\end{figure}

\section{Experiments}
We aim to answer two questions to evaluate our imputation approach: Do the imputed embeddings encode semantic similarity and relatedness information as judged by domain experts? And, can the imputed embeddings be reliably used alongside the original, non-imputed word embeddings?

We use the UMNSRS dataset to answer these questions \citep{pakhomovSemanticSimilarityRelatedness2010}.
It is a collection of medical word-pairs annotated with a relatedness and similarity score by healthcare professionals, such as medical coders and clinicians; some examples are shown in Table~\ref{table:umnsrs-pairs}.
For each word-pair we calculate the cosine similarity between the corresponding word embedding vectors and report the Pearson correlation between these cosine similarities and the human scores.

\begin{table}[h]
\small 
    \begin{tabularx}{\columnwidth}{llrr}
    \toprule
    Term1 & Term2 & Similarity & Relatedness \\
    \midrule
    Acetylcysteine & Adenosine & 256.25 & 586.50 \\
    Anemia & Coumadin & 623.75 & 926.50\\
    Rales & Lasix & 742.00 & 1379.50\\
    Tuberculosis & Hemoptysis & 789.50 & 1338.50\\
    \bottomrule
    \end{tabularx}
    \caption{Examples of UMNSRS word pairs. Scores range from 0 to 1600 (larger = more similar/related).}
    \label{table:umnsrs-pairs}
\end{table}

To obtain additional insight into the performance of the imputation procedure we split the words in the UMNSRS dataset into two groups of roughly the same size: one group of words (\textit{trained}) which we train directly as part of the word embedding training and another group of words (\textit{imputed}) which we obtain via imputation. This split results in three word-pair subsets that contain \textit{imputed/imputed} word pairs, \textit{trained/trained} word pairs, and \textit{imputed/trained} word pairs.  
Note that due to an incomplete overlap of the UMNSRS test vocabulary with both the \gls{mesh} node labels and our word embedding vocabulary we cannot evaluate on every word pair in UMNSRS (see Table \ref{table:UMNSRS_num_test_cases} for more details).
Applying the UMNSRS evaluation to these three groups of word pairs we aim to measure the extent to which the imputation procedure encodes domain-specific semantic information.

For word embedding training we prepare a corpus of 74.4M sentences from open access publications on PubMed (from \url{https://ftp.ncbi.nlm.nih.gov/pub/pmc/oa_bulk/}; accessed on 2021-08-30). 
To simulate the problem of missing words as realistically as possible we then prepare a filtered version of this corpus by removing any sentence containing one of the \textit{imputed} terms (in either singular or plural form). This filtering removes 2.36M of the 74.4M sentences (3.2\%).

We then train 200-dimensional skip-gram word embedding models on both the full and the filtered version of the training corpus. In addition, we also train fastText embeddings \citep{bojanowskiEnrichingWordVectors2017} on both the full and the filtered corpus. For details on the hyper-parameters see Appendix \ref{sec:app:hyperparams_word_embeddings}. Since fastText, which represents words as n-grams of their constituent characters, has been shown to give reasonable embedding vectors for words which are rare or missing in the training corpus it represents a suitable baseline to which we can compare our imputation procedure.

We check that the embedding models (both skip-gram and fastText) trained on the filtered corpus perform roughly on par with those trained on the full corpus when evaluated using the \textit{trained/trained} subset of the UMNSRS test data. 
We also check that the skip-gram model trained on the full corpus performs comparable to the BioWordVec model \citep{zhangBioWordVecImprovingBiomedical2019} across all subsets of UMNSRS. See Appendix \ref{sec:app:add_results_umnsrs} for details.

\gls{lsi} is a means of leveraging the domain space to create \gls{oov} embedding vectors. As a simple alternative baseline,  we directly use the domain space embeddings for the \gls{oov} words. We need to align the domain space onto the semantic space, which we do with a rotation matrix derived from the anchor term embeddings in the two spaces via singular value decomposition.

\subsection{Results}
\begin{figure}[h]
\centering
\begin{subfigure}[b]{\columnwidth}
\centering
\caption{UMNSRS similarity.}
\includegraphics[width=\textwidth]{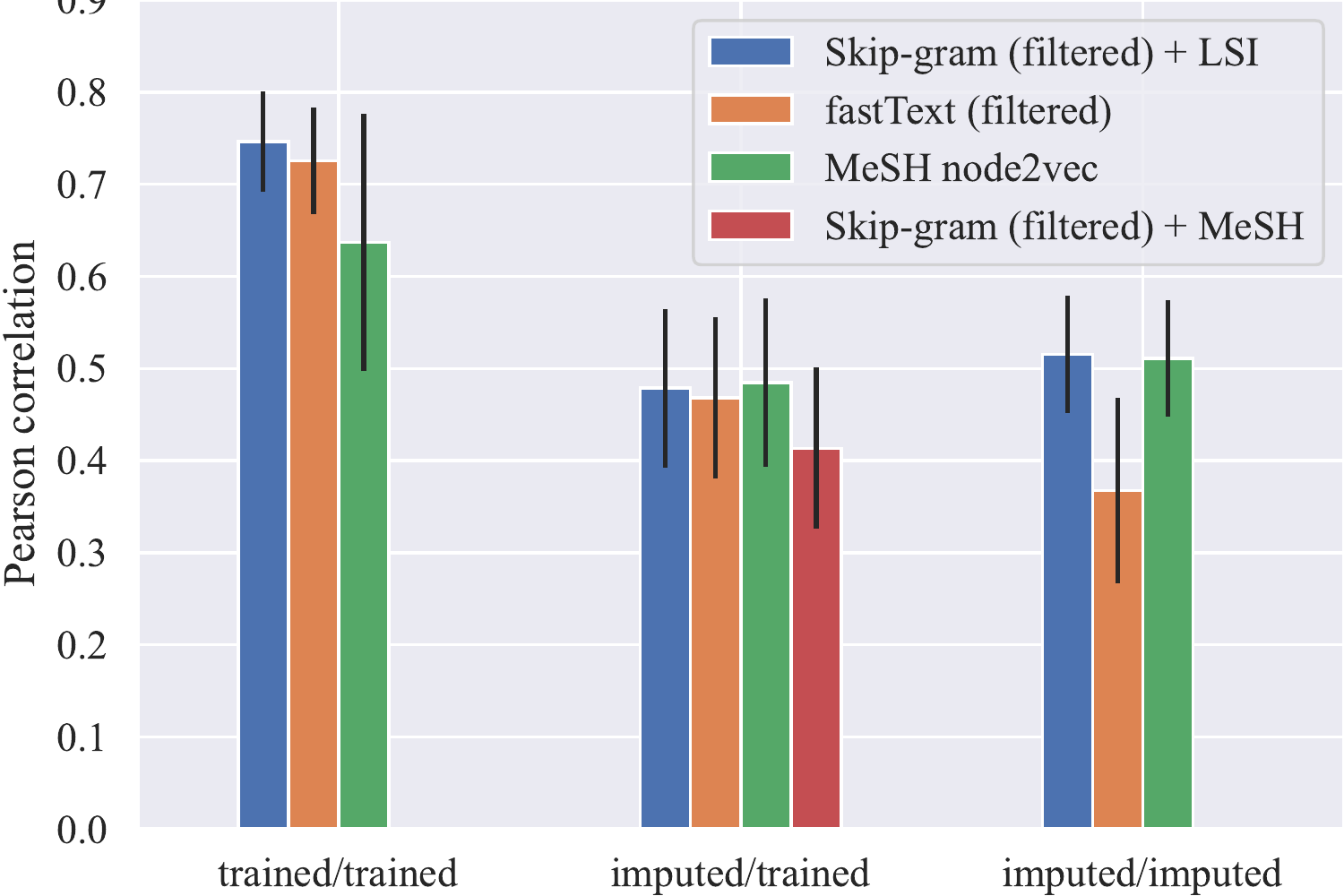}
\label{fig:UMNSRS_sim}
\end{subfigure}
\hfill
\begin{subfigure}[b]{\columnwidth}
\centering
\caption{UMNSRS relatedness.}
\includegraphics[width=\textwidth]{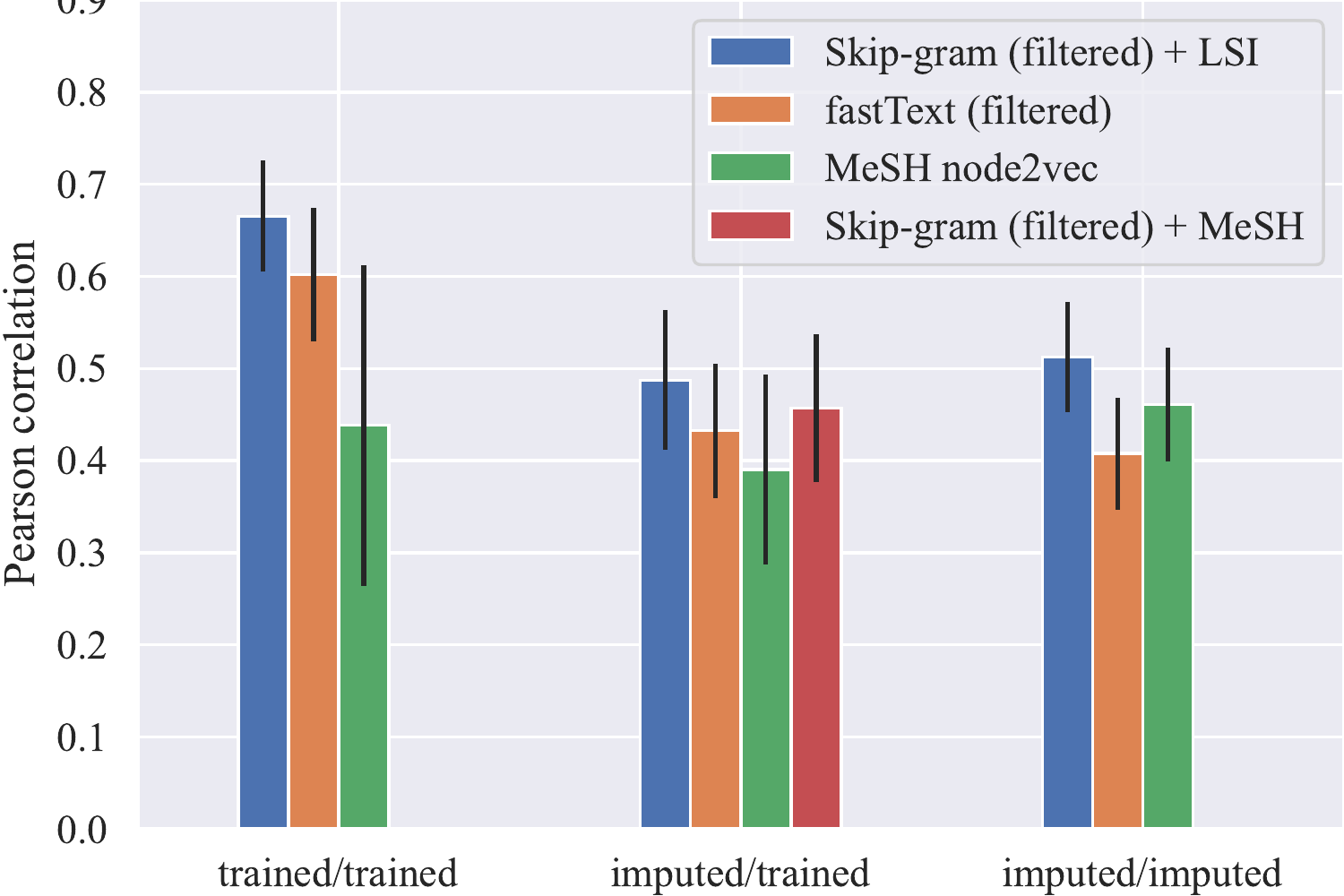}
\label{fig:UMNSRS_rel}
\end{subfigure}
\vspace*{-9mm}
\caption{Correlation with UMNSRS scores.}
\label{fig:UMNSRS}
\end{figure}

The main results are displayed in Fig.~\ref{fig:UMNSRS} which shows the Pearson correlation between cosine similarities and human annotator scores for UMNSRS similarity and relatedness. 
The error bars are standard deviations across 1,000 bootstrap resamples of the test dataset.
From left to right we show results for the \textit{trained/trained}, \textit{imputed/trained}, and \textit{imputed/imputed} subsets. 

We compare two models trained on the filtered corpus (which does not contain any mentions of the \textit{imputed} words): a skip-gram model extended by \gls{lsi} and a fastText model. 
For reference we also show the correlation strengths obtained when directly using the \gls{mesh} node embeddings which form the basis of the imputation. Note that for this last model, the test cases we evaluate are different, since the \gls{mesh} model cannot represent all word pairs in UMNSRS (see appendix \ref{sec:app:add_results_umnsrs} for details). Uncertainties on the \gls{mesh} model are high for the \textit{trained/trained} subset due to the limited overlap of the \gls{mesh} model with the words in the \textit{trained} subset (see Table \ref{table:UMNSRS_num_test_cases}).

In Fig.~\ref{fig:UMNSRS} the \textit{imputed/trained} group also includes the performance of the simple baseline, \textit{Skip-gram (filtered) + MeSH}, formed of a mixture of aligned embeddings. We do not show the performance of this baseline on the other two groups since, by construction, it is identical to that of \textit{Skip-gram (filtered) + LSI} for \textit{trained/trained} and that of \textit{MeSH node2vec} for \textit{imputed/imputed}.

Three things stand out: 
\begin{enumerate}
\item The \gls{lsi}-based model is competitive on novel vocabulary: it performs significantly better than the fastText model on word pairs containing only imputed terms (\textit{imputed/imputed}) and modestly better on mixed word pairs (\textit{imputed/trained}). It also outperforms the simple but surprisingly strong baseline, \textit{Skip-gram (filtered) + MeSH}. 
\item There is a significant difference in Pearson correlation between the different word pair categories. 
Note that the same trend in correlation across word pair categories can be seen in the word embedding model trained on the full corpus without imputation (see Fig.~\ref{fig:UMNSRS_lsi_full}). 
\item The \gls{lsi}-based model obtains better scores than the underlying \gls{mesh} node embeddings across most categories. This proves that the similarity and relatedness information directly encoded in the domain embedding does not limit the similarity and relatedness information encoded in the resulting imputed model.
\end{enumerate}

\subsection{Discussion}
In this paper we use a significantly larger subset of the \gls{mesh} graph compared to related work on \gls{mesh}-based embeddings  \citep{guoMeSHHeading2vecNewMethod2021, zhangBioWordVecImprovingBiomedical2019} by including more than just the topical descriptor nodes. Using a larger graph for the imputation allows us to impute a wider variety of words and evaluate the imputation procedure on a larger subset of UMNSRS. The graph we use for imputation is also much larger than the domain data used in previous work on \gls{lsi} \citep{yaoEnhancingDomainWord2019}. This shows that \gls{lsi} can apply to knowledge graphs and scale to larger domain spaces which is crucial for real-world applications.

We observe that the UMNSRS similarity and relatedness correlations of the \gls{mesh} node embedding models do not constitute an upper bound on the correlations obtained for the imputed word embeddings. This is intuitively plausible since \gls{lsi} combines the global structure of the trained word embedding vectors with the local structure of the domain embeddings. This is in contrast to the original LSI paper in which the domain data alone was sufficient to obtain near perfect scores on the evaluation task and, as such, could have been used directly which obviates the need for \gls{lsi}. This observation reduces the pressure for an optimal knowledge graph and associated embedding, although a systematic search for better subgraphs to use is likely to yield improved imputation results.

It is also of note that most of the trends displayed by the \gls{lsi} model hold for both the similarity and relatedness scores, despite these being distinctly separate concepts. Relatedness is a more general measure of association between two terms whilst similarity is a narrower concept tied to their likeness. This might not be the case if the graph construction had been limited to particular relationship types or if direction of the relations had been retained.

There are noteworthy differences between our experiment and the use cases we envisage for LSI. The words we impute in our experiment are taken from the constituent words of the UMNSRS word pairs rather than being solely defined by training corpus statistics.  This is a necessary limitation of our evaluation methodology. It remains a question for further research to establish ways of evaluating embedding quality on a larger variety of \gls{oov} words and use this for a broader analysis of the performance of \gls{lsi}.

\section{Strengths and weaknesses of \gls{lsi}}
Our experiments highlight several beneficial features of \gls{lsi}. It is largely independent of the nature of the domain data as long as embeddings for the domain entities can be inferred. 
It does not rely on retraining the word embedding and is therefore applicable to cases where retraining is not an option due to limitations in compute or because of lack of access to the training corpus. 
It allows word embeddings to be improved on demand for specific \gls{oov} terms, thus affording a high level of control. In particular, it allows controlled updates of word embeddings in light of new emerging research.

The current challenges we see for \gls{lsi} are driven by limited research in the constituent steps of the imputation pipeline.
Specifically, there is not yet a principled answer for the optimal selection of a subgraph from the full knowledge graph or the optimal choice of node embedding architecture. The answer to these may depend on the domain knowledge graph. 
Also, there are not yet generic solutions for quality control of \gls{lsi}. This problem is likely intrinsically hard since the words which are most interesting for imputation are novel or rare and thus exactly the words for which little data is available.

\section{Conclusion}
In this paper, we show how \gls{lsi} can be used to improve word embedding models for the biomedical domain using domain-specific knowledge graphs.
We use an intrinsic evaluation task to demonstrate that \gls{lsi} can yield good embeddings for domain-specific \acrlong{oov} words.

We significantly extend the work of \citet{yaoEnhancingDomainWord2019} by showing that \gls{lsi} is applicable to scientific text where problems with rare and novel words are particularly acute. \Citet{yaoEnhancingDomainWord2019} assumed a small number of domain entities and a numeric domain data feature matrix. This immediately yields the metric structure required to determine the nearest neighbors and minimum spanning tree graph used in \gls{lsi}. We extend this to a much larger number of domain entities and to domain data which does not have an a priori metric structure but is instead given by a graph structure. We demonstrate that \gls{lsi} can also work with relational domain data thus opening up a broader range of data sources. The metric structure induced by node embeddings trained on a domain knowledge graph provides an equally good starting point for \gls{lsi}. 

This shows that \gls{lsi} is a suitable methodology for controlled updates and improvements of scientific word embedding models based on domain-specific knowledge graphs.

\section{Future work}
We see several fruitful directions for further research on \gls{lsi} and would like to see \gls{lsi} applied to other scientific domains, thereby testing the generalizability of our methodology. This would also provide more insight on how the domain knowledge graph as well as the node embedding architecture impact the imputation results.

The use of automatic methods for creating medical term similarity datasets \citep{schulzLargeScaleDatasets} would facilitate the creation of large-scale test sets. The UMNSRS dataset, along with the other human-annotated, biomedical word pair similarity test sets used in the literature, all consist of fewer than one thousand word pairs \citep{pakhomovCorpusDomainEffects2016, pakhomovSemanticSimilarityRelatedness2010, chiu2018}. The use of larger test sets would remove the aforementioned evaluation limitations. 

Further research could elucidate how to best utilize the full information of the domain knowledge graph in \gls{lsi}. This includes information about node and edge types, as well as literal information such as human-readable node labels and numeric node properties (such as measurement values). 
It also remains to be studied how to optimally choose the anchor terms (to be used in the imputation step) to maximize \gls{lsi} performance. 
Our methodology could also be generalized from \glsentrylong{lsi} to what might be called latent semantic information fusion where domain information is used for incremental updates instead of outright replacement of word embedding vectors.

Finally, \gls{lsi} could also be extended to provide alignment between knowledge graphs and written text
by using the spatial distance between imputed vectors of knowledge graph nodes and trained word embedding vectors as an alignment criterion.

\section*{Acknowledgements}
This paper was supported by the AI Chemist funding (Project ID: 309594) from the Research Council of Norway (RCN).
We thank Shibo Yao for helpful input and for sharing raw data used in \citep{yaoEnhancingDomainWord2019} and Dr.~Zhiyong Lu and Dr.~Yijia Zhang of the National Institute of Health for sharing their word embedding models.
We thank the three anonymous reviewers for their careful reading and helpful comments.

\bibliographystyle{acl_natbib}
\bibliography{custom}

\section*{Appendix}
\subsection{Hyper-parameters for \gls{mesh} node2vec}
\label{sec:app:hyperparams_mesh_n2v}
We train node2vec (\url{https://github.com/thibaudmartinez/node2vec}) embeddings with the hyperparameters shown in Table~\ref{table:mesh_n2v_hyperparams} from a subgraph of \gls{mesh} containing 58,695 nodes and 113,094 edges.
\begin{table}[h]
\begin{tabularx}{\columnwidth}{llr}
\toprule
Hyperparameter         & Variable name & Value \\
\midrule
Training epochs        & epochs        & 50    \\
No. of random walks & n\_walks      & 10    \\
Return parameter       & p             & 0.5   \\
Inout parameter        & q             & 0.5   \\
Context window         & context\_size & 15    \\
Dimension              & dimension     & 200  \\
\bottomrule
\end{tabularx}
\caption{Hyperparameters for MeSH node2vec training}
\label{table:mesh_n2v_hyperparams}
\end{table}

\subsection{Hyper-parameters for word embeddings}
\label{sec:app:hyperparams_word_embeddings}
We use gensim (\url{https://radimrehurek.com/gensim}; version~4.1.2.) for training skipgram and fastText word embedding models with the hyperparameters provided in Table \ref{table:word_embedding_hyperparams}. All other hyperparameters are set to the default values of the gensim implementation. For the skipgram model we use the hyperparameters from \citet{chiuHowTrainGood2016}, which are reported  to be optimal for the biomedical domain. For fastText we are not aware of literature on optimal hyperparameters for the biomedical domain so we use the default values except for the embedding dimension which we set to 200 to ease comparison with the skipgram model. We trained the fastText models for 10 epochs but found that the performance of the fastText model on UMNSRS saturates after epoch~1. We use the fastText model after the first epoch for the remainder of our experiments and analysis. 

\begin{table}[h]
\begin{tabularx}{\columnwidth}{lrr}
\toprule
Variable name~~~~ & ~~~~fastText & ~~~~skipgram \\
\midrule
epochs & 1 & 10 \\
negative & 5 & 10 \\
vector\_size & 200 & 200 \\
alpha & 0.025 & 0.05 \\
sample & 1E-03 & 1E-04 \\
window & 20 & 30 \\
\bottomrule
\end{tabularx}
\caption{Hyperparameters for skipgram and fastText training. See the gensim documentation for the definition of the hyperparameters.}
\label{table:word_embedding_hyperparams}
\end{table}

\begin{figure*}
\centering
\begin{subfigure}[b]{\columnwidth}
\centering
\caption{UMNSRS similarity.}
\includegraphics[width=\textwidth]{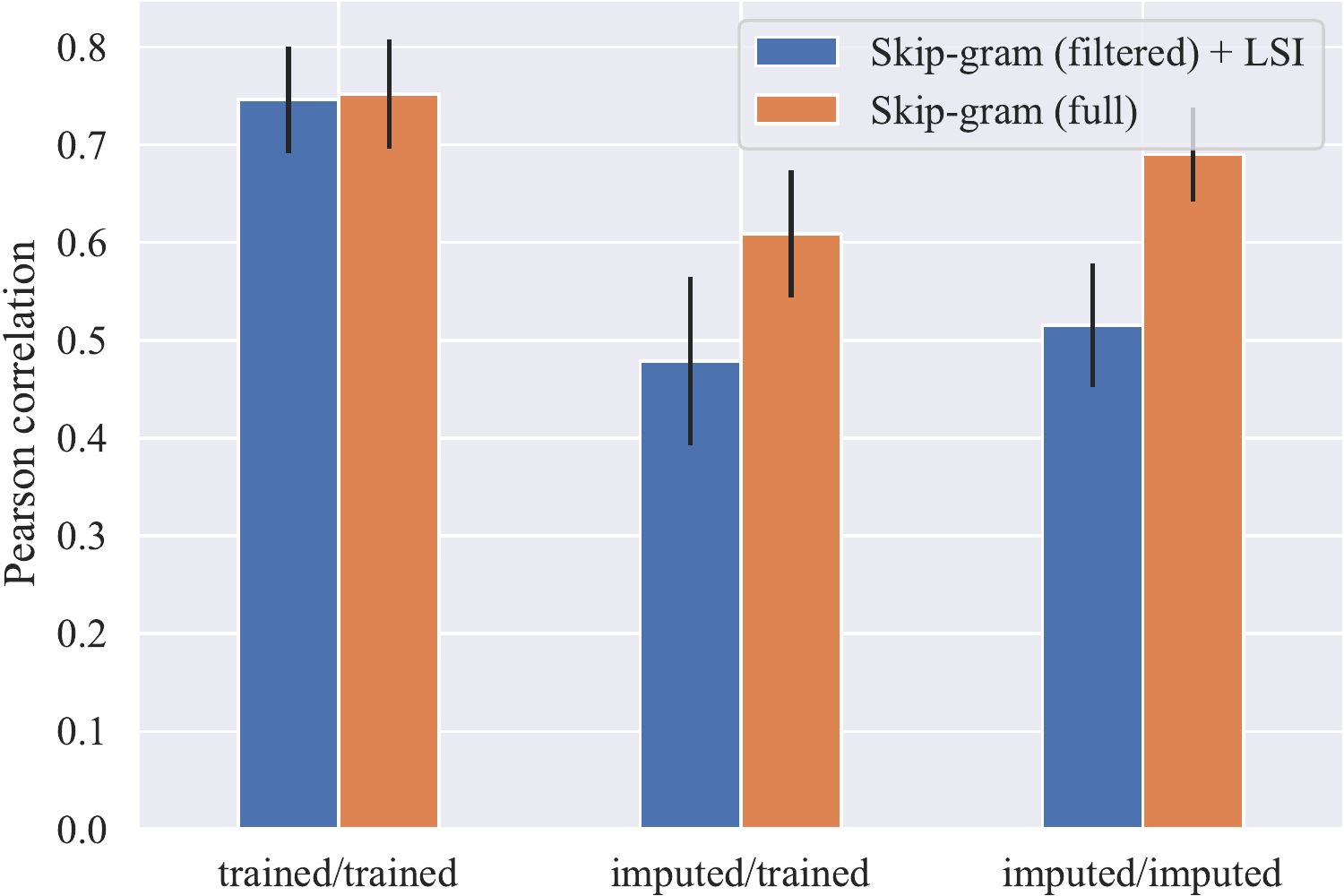}
\label{fig:UMNSRS_sim_lsi_full}
\end{subfigure}
\hfill
\begin{subfigure}[b]{\columnwidth}
\centering
\caption{UMNSRS relatedness.}
\includegraphics[width=\textwidth]{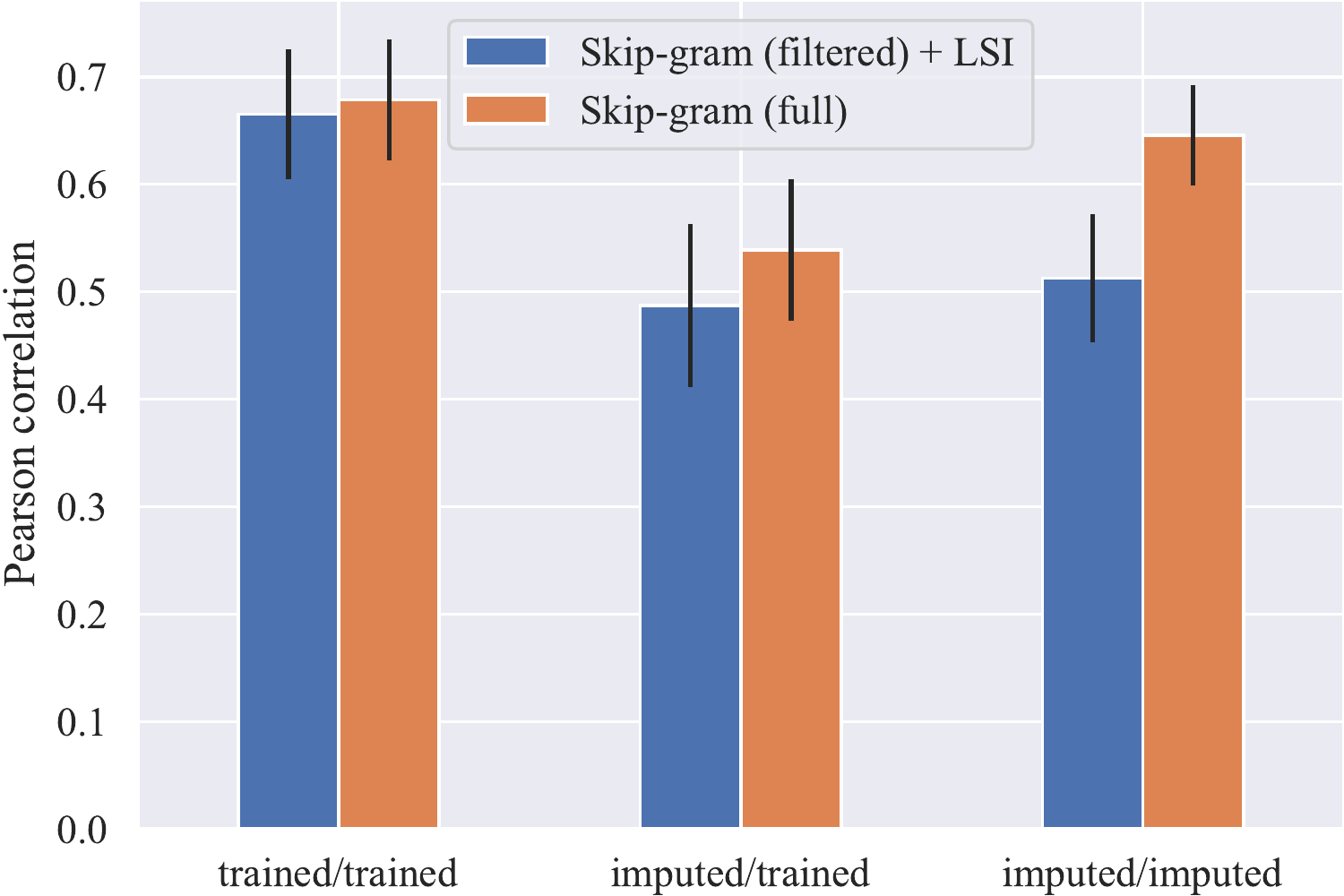}
\label{fig:UMNSRS_rel_lsi_full}
\end{subfigure}
\vspace*{-3mm}
\caption{UMNSRS correlations for skipgram models.}
\vspace*{12mm}
\label{fig:UMNSRS_lsi_full}

\begin{subfigure}[b]{\columnwidth}
\centering
\caption{UMNSRS similarity.}
\includegraphics[width=\textwidth]{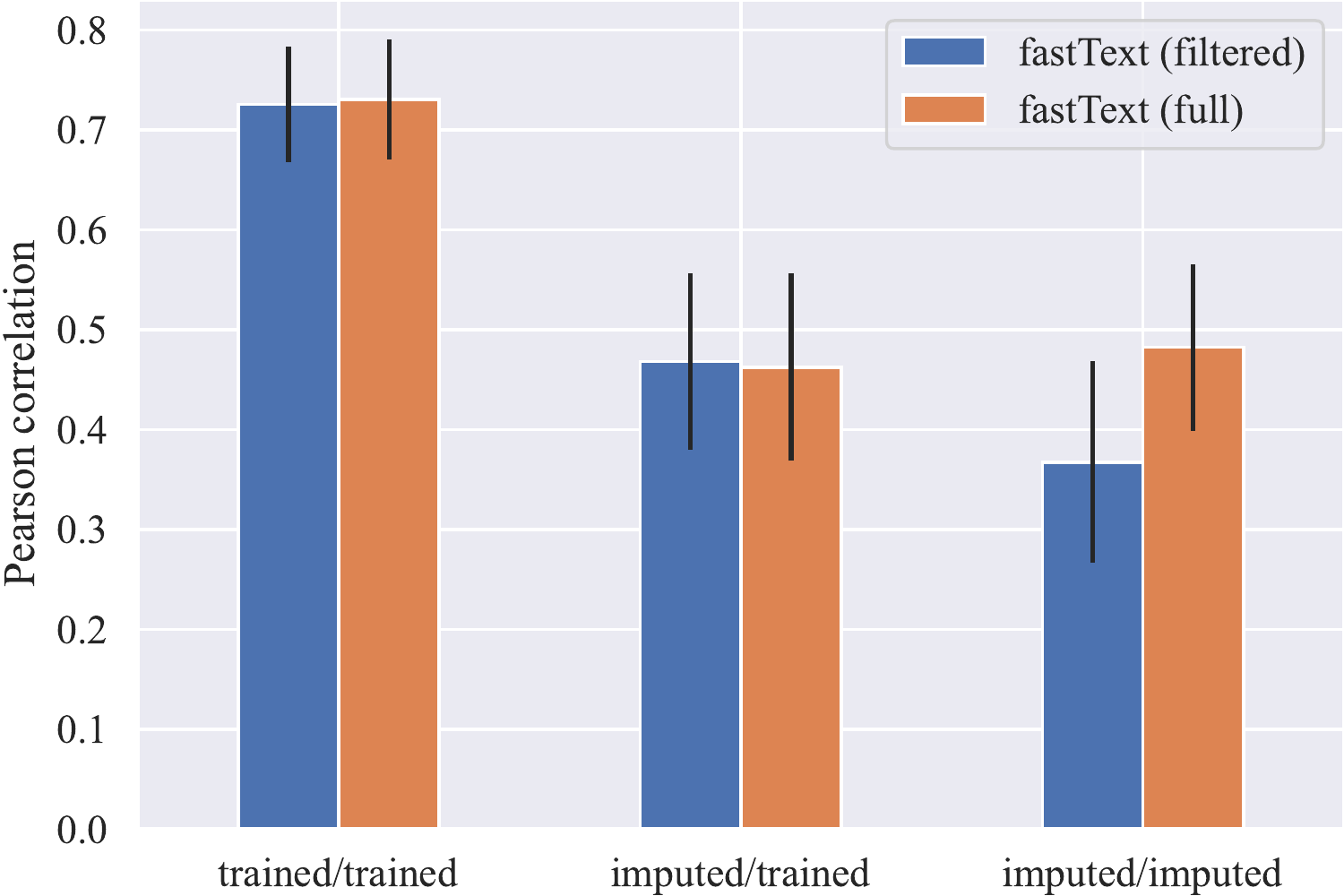}
\label{fig:UMNSRS_sim_fasttext_full}
\end{subfigure}
\hfill
\begin{subfigure}[b]{\columnwidth}
\centering
\caption{UMNSRS relatedness.}
\includegraphics[width=\textwidth]{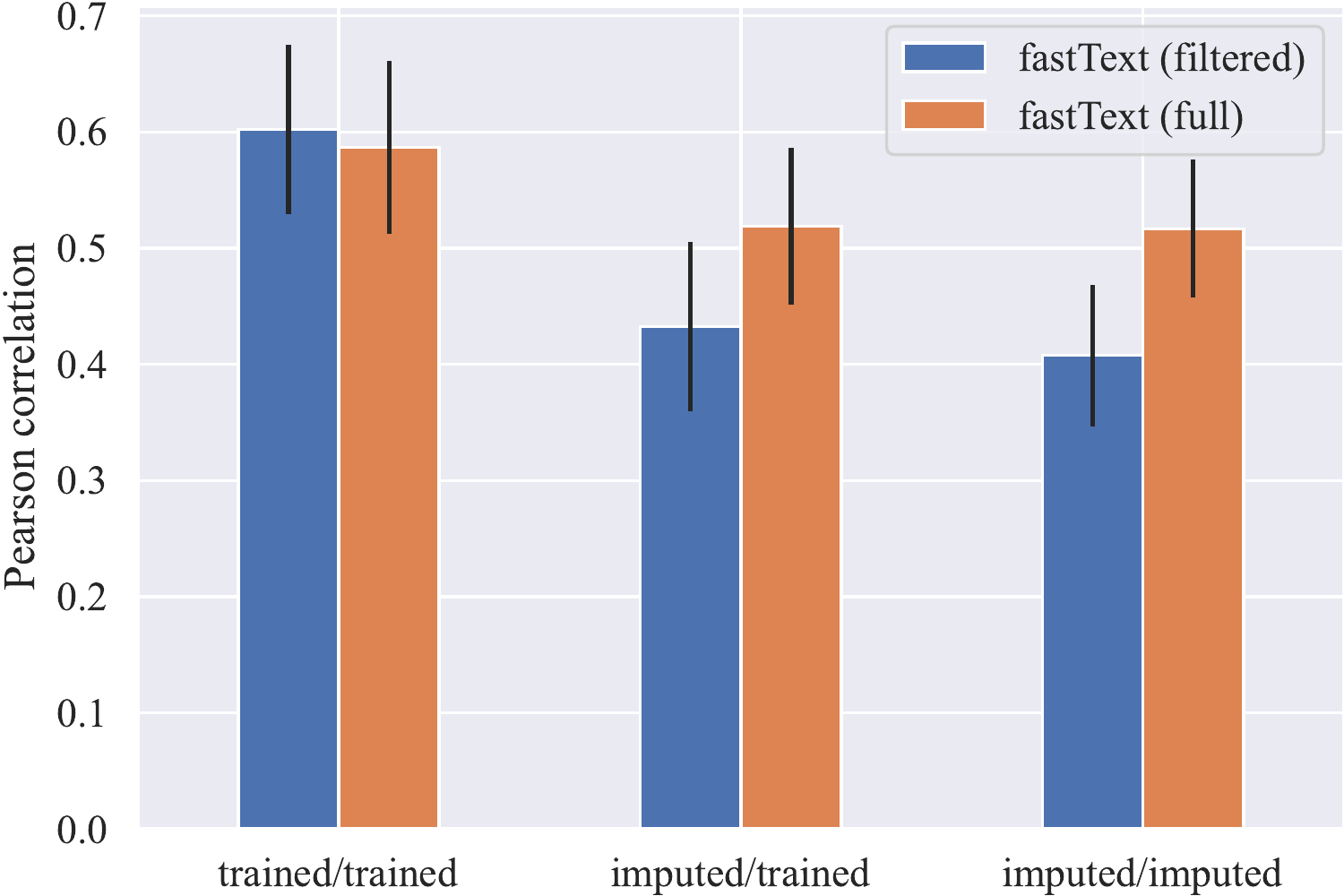}
\label{fig:UMNSRS_rel_fasttext_full}
\end{subfigure}
\vspace*{-3mm}
\caption{UMNSRS correlations for fastText models.}
\vspace*{12mm}
\label{fig:UMNSRS_fasttext_full}

\begin{subfigure}[b]{\columnwidth}
\centering
\caption{UMNSRS similarity.}
\includegraphics[width=\textwidth]{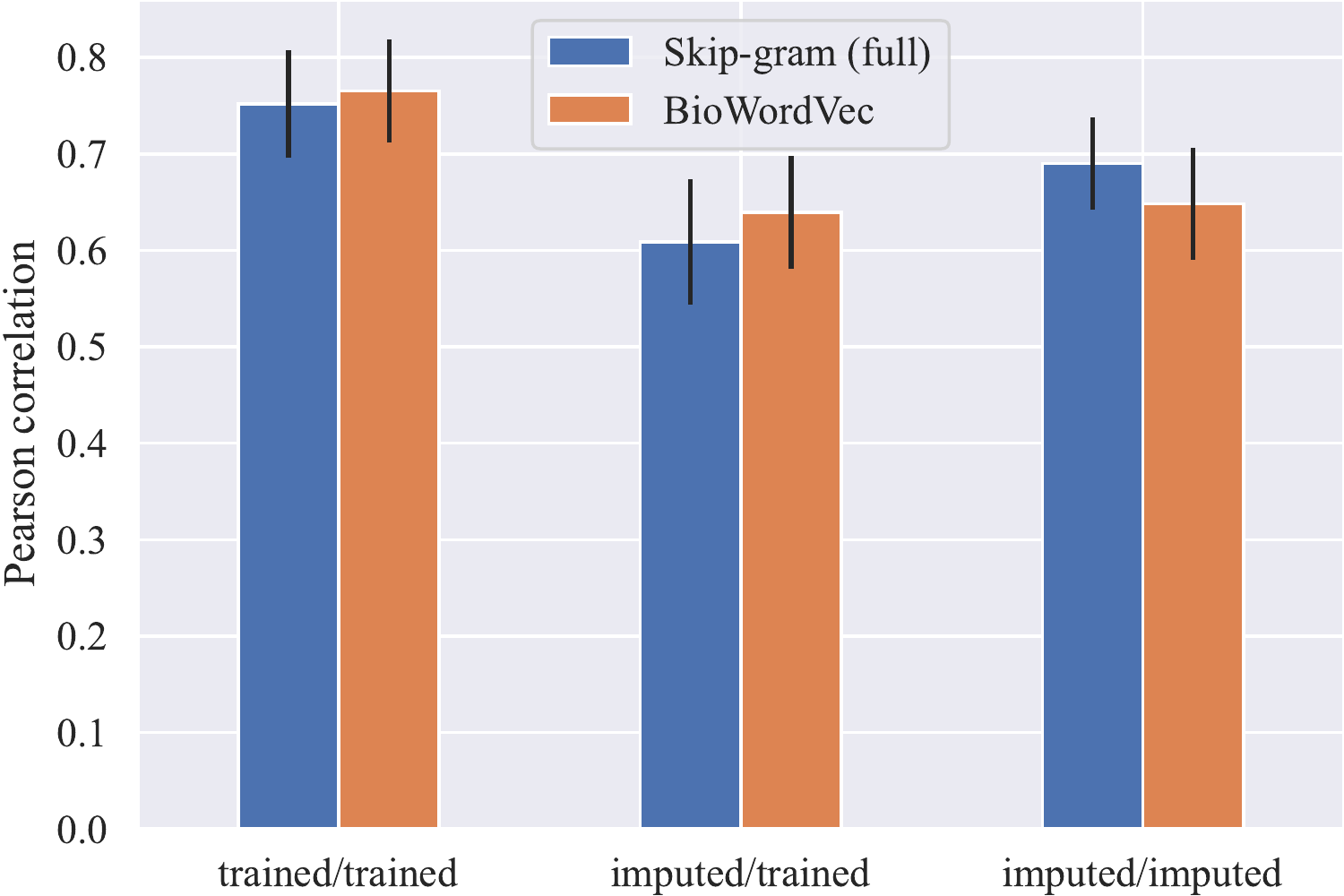}
\label{fig:UMNSRS_sim_biowordvec_full}
\end{subfigure}
\hfill
\begin{subfigure}[b]{\columnwidth}
\centering
\caption{UMNSRS relatedness.}
\includegraphics[width=\textwidth]{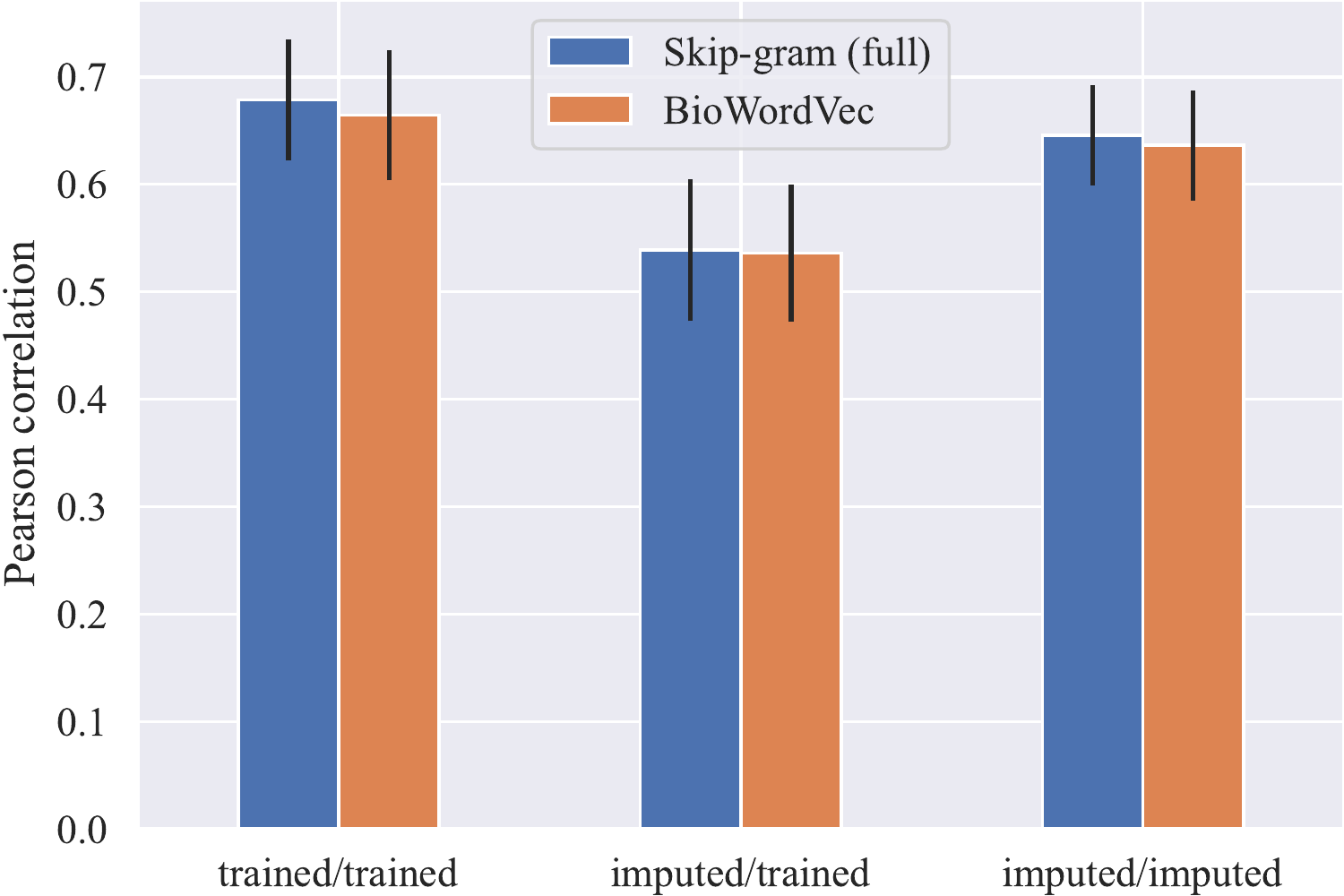}
\label{fig:UMNSRS_rel_biowordvec_full}
\end{subfigure}
\vspace*{-3mm}
\caption{UMNSRS correlations for BioWordVec.}
\label{fig:UMNSRS_biowordvec_full}
\end{figure*}

\begin{table*}[h]
\begin{tabularx}{\textwidth}{lrrrrrr}
\toprule
& \multicolumn{3}{c}{UMNSRS relatedness} & \multicolumn{3}{c}{UMNSRS similarity} \\
Model & ~~~~~~~~\begin{tabular}[c]{@{}l@{}}trained/\\ trained\end{tabular} & \begin{tabular}[c]{@{}l@{}}imputed/\\ trained\end{tabular} & \begin{tabular}[c]{@{}l@{}}imputed/\\ imputed\end{tabular} & ~~~~~~~~\begin{tabular}[c]{@{}l@{}}trained/\\ trained\end{tabular} & \begin{tabular}[c]{@{}l@{}}imputed/\\ trained\end{tabular} & \begin{tabular}[c]{@{}l@{}}imputed/\\ imputed\end{tabular} \\
\midrule 
MeSH node2vec & 28 & 70 & 133 & 30 & 72 & 135 \\ 
all other models & 83 & 99 & 124 & 84 & 101 & 126 \\
\bottomrule 
\end{tabularx}
\caption{Number of test cases per model and test set split for UMNSRS evaluation.}
\label{table:UMNSRS_num_test_cases}
\end{table*}

\subsection{Details on the UMNSRS evaluation}
\label{sec:app:add_results_umnsrs}
Table \ref{table:UMNSRS_num_test_cases} shows the number of test cases per model and UMNSRS test data split. All models have been evaluated on the same subsets of UMNSRS except for the \gls{mesh} node embeddings model where limited overlap with the UMNSRS test vocabulary prevents us from evaluating on exactly the same subsets.

The embedding models (both skip-gram and fastText) trained on the filtered corpus perform roughly on par with those trained on the full corpus when evaluated using the \textit{trained/trained} subset of the UMNSRS test data (see Fig.~\ref{fig:UMNSRS_lsi_full} and \ref{fig:UMNSRS_fasttext_full}). When comparing the performance of the filtered skipgram model + \gls{lsi} to the full skipgram model on the subset of test data involving imputed words (\textit{imputed/trained} and \textit{imputed/imputed}) the full model outperforms \gls{lsi} (see Fig.~\ref{fig:UMNSRS_lsi_full}). This suggests that, if training text for the \gls{oov} words were available, we should make use of it. Similarly, and as expected, when comparing the performance of the filtered and full fastText models on the subset of test data involving imputed words (\textit{imputed/trained} and \textit{imputed/imputed}) the full model again outperforms the filtered model (see Fig.~\ref{fig:UMNSRS_fasttext_full}).

As a sanity check, we also compare the skip-gram model trained on the full corpus to BioWordVec, a recent state-of-the-art word embedding model for the biomedical domain \citep{zhangBioWordVecImprovingBiomedical2019} and find similar performance across all subsets of UMNSRS (see Fig.~\ref{fig:UMNSRS_biowordvec_full}).
\end{document}